\documentclass{ecai2008}

\usepackage{fixltx2e}
\usepackage[utf8]{inputenc}
\usepackage[T1]{fontenc}
\usepackage{times}
\usepackage{graphicx}
\usepackage{latexsym}
\usepackage[english]{babel}
\usepackage{url}
\usepackage{textcomp}
\usepackage{microtype}
\usepackage[babel]{csquotes}
\usepackage{amsmath}
\usepackage{amssymb}
\usepackage{upgreek}
\usepackage{ellipsis}
\usepackage{booktabs}

\begin{document}

\title{Extracting Features from Ratings:\\The Role of Factor Models}

\author{Joachim Selke \and Wolf-Tilo Balke\institute{Institut für Informationssysteme, Technische Universität Braunschweig, Germany}}

\maketitle
\bibliographystyle{ecai2010}

\begin{abstract}
	Performing effective preference-based data retrieval requires
	detailed and preferentially meaningful structurized information
	about the current user as well as the items under consideration.
	A common problem is that representations of items often only
	consist of mere technical attributes, which do not resemble human perception.
	This is particularly true for integral items such as movies or songs.
	It is often claimed that meaningful item features could be extracted
	from collaborative rating data, which is becoming available
	through social networking services. However, there is only anecdotal evidence
	supporting this claim; but if it is true,
	the extracted information could very valuable
	for preference-based data retrieval.
	In this paper, we propose a methodology to systematically
	check this common claim. We performed a preliminary
	investigation on a large collection of movie ratings
	and present initial evidence.
\end{abstract}

\section{INTRODUCTION}

Recommender systems \cite{RecSys, RecSysCACM} are one of the most prominent applications
of preference handling technology \cite{PrefHandling} and a highly active area of research.
In particular, fueled by the Netflix competition and its one million dollar prize money \cite{Netflix},
research on collaborative recommendation techniques \cite{CFRecSys} has recently made significant advances,
most notably through the introduction of \emph{factor models} \cite{KorenBellVolinskyFact, UnifiedFact}.

In collaborative recommender systems, \emph{users} repeatedly express their preferences for \emph{items},
which usually is done by giving explicit \emph{ratings} on some predefined numerical scale. This data
can be modeled using a \emph{rating matrix,} whose rows correspond to items, columns to users, and entries
to ratings. Typically, ratings matrices are very sparse, that is, only a small fraction of all possible
ratings have actually been observed. Personalized recommendations are generated by predicting
unobserved ratings from the available data and, for each user, selecting those items considered to be most appealing.

Most state-of-the-art collaborative recommendation methods---including the winner of the Netflix Prize---are
based on factor models, which are known to yield much more accurate predictions than traditional neighborhood-based methods
\cite{KorenTemporal, KorenFactNgbr, UnifiedFact, NNMFTakacs, MMMF}.
In factor models, each user and each item is represented
by a vector in some shared real coordinate space. The vectors are chosen such that each observed
rating is closely approximated by the dot product of the corresponding item and user vectors. The selection
of coordinates usually is formalized as an optimization problem. Predictions for unobserved ratings
are generated by computing the respective scalar products. Equivalently, this approach can be seen
as a factorization of the rating matrix into the product of an item matrix (whose rows are the item vectors)
and a user matrix (whose columns are the user vectors). 

The success of factor models is usually attributed to the intuition that the coordinate space
used to represent items and users actually is a \emph{latent feature space}. That is, its dimensions
capture the items' perceptual properties as well as the users' preference judgments regarding
these properties. For example, when items are movies, the individual dimensions are generally thought
to measure (more of less) \enquote{obvious} features such as horror vs. romance, the level of sophistication,
or orientation towards adults. For users, each coordinate is thought to describe the relative degree of importance
attached to the respective dimension. This understanding of factor models can be found throughout the literature,
for example, in \cite{Netflix, KorenFactNgbr, KorenBellVolinskyFact, FunkSVD, NNMFTakacs}.

Although it is intuively appealing, to our knowledge, the correspondence to features has never been
systematically proven, but is only reported anecdotically. For example, Koren et al. \cite{KorenBellVolinskyFact} performed
a factorization on the Netflix movie data set and manually interpreted the first two coordinates
for selected movies as follows:
\begin{quote}
	Someone familiar with the movies shown can see clear meaning in the latent factors. The first
	factor has on one side lowbrow comedies and horror movies,
	aimed at a male or adolescent audience,
	while the other side contains drama or comedy with serious undertones and strong
	female leads. The second factorization
	axis has independent, critically acclaimed, quirky films
	on the top, and on the bottom, mainstream formulaic films.
\end{quote}
Further evidence has been provided by Takács et al. \cite{NNMFTakacs}.
After performing a factorization of the Netflix data set, they manually assigned labels
to individual dimensions of their coordinate space, such as \emph{Legendary,}
\emph{Typical for men,} \emph{Romantic,} and \emph{NOT Monty Python}.

In this paper, we propose a \emph{systematic} method for studying the coordinate spaces derived from factor models
and apply it the MovieLens\,10M data set, a large real-world collection of movie ratings.
The main contribution of our work consists in laying important groundwork,
on which further research in recommender systems and preference handling can be build.
In particular, we see two concrete directions for future work:

\begin{itemize}
	\item
		First, knowing what kind of semantic information is extracted by factor models---and how
		it is represented in coordinate spaces---will enable a deeper understanding of
		these methods. Ultimately, these findings may lead to a more systematic development
		and refinement of recommender systems. In particular, a systematic
		assessment of semantic structures provides an additional way of evaluating the effectiveness
		of factor-based recommenders. This would perfectly complement
		traditional evaluation methods \cite{RecSysEval}, which focus on predictive accuracy.
	\item
		Second, we believe that factor models might be a powerful tool for automatically extracting
		meaningful descriptions of otherwise hard-to-describe items such as movies or songs---particularly,
		essential features of movies cannot be characterized at all by purely technical features such as runtime,
		language, or release date.\footnote{A complementary approach to closing this \emph{semantic gap}
		is content-based image and video retrieval \cite{ES03}.}
		But given a coordinate representation of movies that matches human perception,
		the full machinery developed in preference handling research can be applied \cite{PrefHandling, PrefLearning}.
		For example, clustering techniques can give user an initial high-level impression of the available items,
		item rankings can be learnt from ordinal preference statements \cite{RankSVM} or utilities \cite{BRV09}, and
		the best items can be retrieved by means of Top-k algorithms \cite{IBS08}. 
\end{itemize}

Since our primary research interest lies in applying preference-based retrieval techniques
to item collections, in this paper we will concentrate on evaluating the semantic structures
contained in the item matrix $A$. Performing a similar analysis of the user matrix $B$ may require
entirely different methods.

The paper is structured as follows: After introducing notation and reviewing
the most important factor models, we develop general guidelines on how to evaluate
coordinate spaces for semantic information. Then, we illustrate how to apply
these guidelines to the evaluation of factor spaces generated from movie rating data
and perform experiments on the MovieLens\,10M data set.

\section{PRELIMINARIES}

In the following, we use the variables $i$ and $j$ to identify items,
whereas $u$ and $v$ denote users. We are dealing with ratings given to $I$ items
by $U$ users. Let $R = (r_{i, u}) \in \{\mathbb{R} \cup \emptyset\}^{I \times U}$ be the
corresponding rating matrix, where $r_{i, u} = \emptyset$, if item $i$ has not been rated by user $u$;
otherwise, $r_{i, u}$ expresses the strength of user $u$'s preference for item $i$.
Ratings are usually limited to a fixed integer scale (for example, one to ten stars).
Moreover, $\mathcal{R} = \bigl\{(i, u)\,|\,r_{i, u} \neq \emptyset\bigr\}$ is
the set of all item--user pairs for which ratings are known. Let $n$ be the total number
of ratings observed (the cardinality of $\mathcal{R}$). Typically, $n$ is very small compared
to the number of possible ratings $I \cdot U$ (for example, in the Netflix data set it is $\frac{n}{I \cdot U} \approx 1.4\%$).

Given some target dimensionality $d$, the basic idea underlying factor models is
to find matrices $A = (a_{i, r}) \in \mathbb{R}^{I \times d}$ and $B = (b_{r, u}) \in \mathbb{R}^{d \times U}$
such that their product $\hat{R} = A \cdot B$ closely resembles $R$ on all known entries.
To quantify this notion of \enquote{close resemblance,} the sum of squared errors (SSE)
is popularly chosen. The SSE difference between the rating matrix $R$ and its estimation $\hat{R} = (\hat{r}_{i, u})$
is defined as
\begin{displaymath}
	\text{SSE}\bigl(R, \hat{R}\bigr) = \sum_{(i, u) \in \mathcal{R}} \bigl(r_{i, u} - \hat{r}_{i, u}\bigr)^2.
\end{displaymath}
Factor models are typically formulated as optimization problems over $A$ and $B$, in which the SSE (or some
other measure) is to be minimized.

Probably the most popular factor model is Brandyn Webb's regularized SVD model \cite{KorenBellVolinskyFact, FunkSVD},
in which $A$ and $B$ are defined as the solution of the least squares problem
\begin{displaymath}
	\min_{A, B}\quad\text{SSE}\bigl(R, A \cdot B\bigr) + \lambda\hspace{-0.3em} \sum_{(i, u) \in \mathcal{R}} \sum_{r = 1}^d \left(a_{i, r}^2 + b_{r, u}^2\right)\text{.}
\end{displaymath}
Here, $\lambda \geq 0$ is a regularization constant used to avoid overfitting.

More advanced versions of the SVD model exclude systematic rating deviations from the factorization and model them explicitly
using new variables. Bell and Koren \cite{SVDB} propose to estimate rating $r_{i, u}$ by
\begin{displaymath}
	\hat{r}_{i, u} = \mu + \delta_i + \delta_u + \sum_{r = 1}^d a_{i, r} b_{r, u}\text{,}
\end{displaymath}
where the constant $\mu$ denotes the mean of all observed ratings; $\delta_i$ and $\delta_u$ are $I + U$ new model parameters expressing
systematic item and user deviations from $\mu$. Again, the parameters are chosen according to a regularized
least squares problem:
\begin{displaymath}
	\min_{A, B, \delta_\star}\quad\text{SSE}\bigl(R, \hat{R}\bigr) + \lambda\hspace{-0.3em} \sum_{(i, u) \in \mathcal{R}} \left(\sum_{r = 1}^d \left(a_{i, r}^2 + b_{r, u}^2\right) + \delta_i + \delta_u\right)\text{.}
\end{displaymath}
The rationale underlying this approach---which we refer to as $\delta$-SVD in the following---is
that the removal of item- and user-specific general trends from the factorization
allows to focus on more sophisticated rating patterns.

The third basic factor model being relevant to our work performs a non-negative factorization of the rating matrix \cite{NNMFTakacs}.
It is identical to the regularized SVD model up to the additional constraint that all entries of $A$ and $B$
must be non-negative. Extending this model by explicit item and users deviations is not reasonable since this
would require negative entries in $A$ and $B$ to approximate $R$ close enough. The non-negative matrix factorization
model aims at creating a coordinate space in which effects of different dimensions on the estimated ratings
cannot cancel out each other. Henceforth, we refer to this model as NNMF.

\section{EVALUATING COORDINATE SPACES}

Given an item--feature matrix $A \in \mathbb{R}^{I \times d}$ generated by some factor model,
how can we determine whether the items' coordinates in this $d$-dimensional space resemble
a \enquote{semantically meaningful} pattern? The most straightforward approach consists in
extending and systematizing the casual investigations described in the introduction.
This could easily be done by presenting the item coordinate space to a number of different
people and asking them to label its dimensions. The correspondence between the generated item coordinates
and human perception could, for example, be done by measuring the degree of consensus among people
or the average time needed to come up with adequate labels.

Although this kind of investigation seems very reasonable,
it contains some severe flaws, which cannot be fixed by careful study design:
\begin{enumerate}
	\item
		The dimensionality chosen in most applications of factor models typically ranges between
		$d = 10$ and $d = 100$. A comprehensive analysis of the resulting
		data sets would require the users to comprehend high-dimensional spaces,
		which is impossible even when using advanced visualization techniques.
	\item
		Due to hindsight bias, given enough time, users will be able to assign a fitting label to almost
		any dimension of the coordinate space. Chances are good that this effect accounts
		for rather questionable labels such as \emph{NOT Monty Python}.
	\item
		By using free association to name dimensions, the collection of resulting labels tend
		to show a high variability and reflect individual differences between users.
		To produce statistically significant results, either the sample size must be extended
		(which requires more study participants and results in higher costs), or the variability must be reduced,
		for example, by training participants to use an established domain-specific vocabulary
		to articulate the semantic properties they recognize in the data (which also increases time and effort).
	\item
		Typically, there are many near-optimal solutions to the above mentioned optimization problems,
		which can be transformed into one another by rotation of the coordinate axes. This is because,
		for any invertible matrix $M \in \mathbb{R}^d$, the solution pairs $(A, B)$ and
		$(AM, M^{-1}B)$ produce the same SSE. Although regularization usually enforces the theoretical existence of
		a unique optimal solution pair, in practice the enormous problem size often
		allows only finding one of the many near-optimal solutions. Consequently, the direction of the coordinate axes
		is completely arbitrary, which makes the task of assigning labels a hopeless undertaking.
\end{enumerate}

\subsection{Some Guidelines}

In this section, we devise a set of guidelines on which to base
more appropriate approaches to the analysis of coordinate spaces.

\begin{itemize}
	\item
		In the view of problems
		(1) and (4), we recommend to avoid any direct human interaction with \emph{item coordinates.}
		Instead, human input should concentrate on describing \emph{item properties,} which in turn are
		related to coordinates as well as compared by algorithmic means.
	\item
		The only effective way
		to eliminate hindsight bias (2) is collecting feedback on items before generating and
		presenting any information extracted by the factor models under consideration.
	\item
		To resolve problem (3), we primarily recommend to adapt a domain-specific vocabulary to allow a
 		structurized description of items. For example, to characterize music, the rich vocubulary
		developed by allmusic\footnote{\url{http://www.allmusic.com}} seems appropriate; amongst others,
		it includes very detailed information about genres, styles, moods, and connections between artists.
		Since this kind of semantic information can be (or already have been) provided by a small number of experts and
		usually is little prone to debate, it is easy to assemble and work with. In later stages of
		analysis, unrestricted user feedback may be included to reveal the position and extent of
		more fine-grained and rather subjective concepts in the coordinate space.
\end{itemize}

We also propose to apply a standardization procedure to the generated coordinate space.
This is for the following reasons: First, recall that, for any invertible matrix
$M \in \mathbb{R}^d$, the solution pairs $(A, B)$ and $(AM, M^{-1}B)$ are equivalent;
to enable comparisons between different factor models and even different runs of
the same optimization algorithms, we need to define one solution pair as the standard representation.
Second, to enable a better separation of different effects in the data, the axes of the item (and user) coordinate space
should be chosen to be orthogonal. Moreover, axes should be ordered according to their relative importance
(measured by the variance of data along each axis); that is, the first dimension should be assigned to the most
important axis.

The perfect tool for matching these requirements is the singular value decomposition, a well-known matrix factorization technique
from linear algebra, which inspired the SVD factor model. It is based on the fact that, for any rank-$d$ matrix $X \in \mathbb{I \times U}$,
there is a column-orthonormal matrix $U \in \mathbb{R}^{I \times d}$, a diagonal matrix $S \in \mathbb{R}^{d \times d}$,
and a row-orthonormal matrix $V \in \mathbb{d \times U}$ such that $X = USV$. By reordering rows and columns,
$S$ can be chosen such that its diagonal elements are ordered by increasing magnitude. Moreover, the diagonal matrix $S$
can be eliminated from this factorization by setting $X = U'V'$, where $U' = U S^{\frac{1}{2}}$ and $V' = U S^{\frac{1}{2}}$.
The matrices $U'$ and $V'$ are unique if all diagonal elements of $S$ have been mutually different.

In our setting, we will apply the singular value decomposition to transform the product $X = A \cdot B$ into
a new product $A' \cdot B'$ as just described. Since rating data tends to be very \enquote{noisy,} we can safely assume
that $(A', B')$ is a unique representation of $(A, B)$; we did not encounter any counterexamples during our
experiments on large real-world rating data. Moreover, any equivalent pair $(AM, M^{-1}B)$ also gets
transformed into $(A', B')$, which we define as the corresponding standard representation.
It can be computed efficiently using the product decomposition algorithm proposed in \cite[Sec.\,3]{ProductSVD}.

\subsection{Use Case: Movie Ratings}

Based on these guidelines, we now present a concrete method for performing 
a basic evaluation of coordinate spaces generated from movie ratings.
Our focus rests on immediate applicability, so we relate the item coordinates
to reference data that is already available.

The reference source for all kinds of movie-related information is IMDb,
the Internet Movie Database\footnote{\url{http://www.imdb.com}}, which currently
covers about 1.6 million titles. Most of IMDb's data has been created with the
help of its users. Therefore, a large proportion of the available content can freely
be downloaded and used for non-commercial purposes\footnote{\url{http://www.imdb.com/interfaces\#plain}}.
Based on this comprehensive data, one should be able to cross-reference any collection of movie ratings with IMDb.

For the semantic evaluations we are going to perform, the following attributes of titles
may prove helpful: genres, certifications (e.g., USA:PG for \emph{parental guidance suggested}),
year of release, and plot keywords. To illustrate the general procedure,
we will only exploit genre information in this paper. Extendig our method to
other types of semantic information is straightforward. Checking the correspondence between
genres and item coordinates also makes up a good first test of whether
at least some basic semantic properties of movies are represented in coordinate spaces, which
is exactly the purpose of the current work.

IMDb recognizes 28 different genres, from \emph{Action} to \emph{Western,} where each movie may belong
to multiple genres. The assignment of genres is done by IMDb's expert staff in cooperation
with IMDb users. To enforce consistency, this process is based upon a collection of publicly available
guidelines\footnote{\url{http://www.imdb.com/updates/guide/genres}}. Therefore, this data source
matches the requirements developed in the previous section.

To analyze whether the distribution of genres in coordinate space displays any
significant pattern, we turn to established classification algorithms, which explicitly
have been designed to exploit any relevant patterns in the data if there are any.
In particular, we propose to measure the degree of adherence to a pattern by
the classification accuracy shown by these algorithms when predicting the genre of movies
based on their coordinates. In essence, we transform our analysis into
a sequence of binary classification problems (one for each genre), which enables us to build on solid grounds.
Following the common methodology, we use cross-validation; that is, accuracy is measured
on a data set, which is independent of the one used to train the classifier.
By applying proven techniques to counter overfitting, our approach also overcomes any possible problems
related to hindsight bias.

For a start, we selected two popular classification algorithms, which are able to detect
different kinds of patterns in the data:
support vector machines and kNN-classifiers.

Support vector machines will be used in two different flavors: first, using a linear kernel (refered to as SVM-lin),
and second, using a Gaussian radial basis function kernel (SVM-RBF). Linear support vector machines
will show a high classification accuracy if most movies of the respective genre are grouped at one
side of the data set, which can be separated from all remaining movies by a hyperplane. For example, this can be used to disprove
the hypothesis that there exists a direction in the coordinate space along which, say, the amount of action,
increases monotonically. In contrast, the SVM-RBF classifier detects whether groups of movies with the same genre
tend to be located in close vincinity.

kNN-classifiers perform well if the distance between movies having the same genre typically is smaller
than the distance to movies not having this genre. Therefore, they can be used to check whether genres form
spatially separated patterns in coordinate space. Since factor models are not based on a notion of proximity,
it is not clear what measure of distance suits factor models best. We will try out the following four measures:
Euclidean distance, standardized Euclidean distance (where, to ensure equally weighted dimensions, coordinate values
are divided by the standard deviation of the data with respect each dimension), negative scalar product
(which essentially adapts the method of rating prediction to measure distance), and cosine similarity
(which is monotonically related to the angle between two vectors).

To evaluate the true benefit of coordinate spaces generated from factor models,
we propose the following baseline, which is derived from traditional neighborhood-based
recommendation methods \cite{ItemItemSim} and constructed as follows:
First, for any items $i$ and $j$, we compute their Pearson correlation coefficient
\begin{displaymath}
	\varrho_{i, j} = \frac{\sum_{u \in \mathcal{R}_{i, j}} (r_{i, u} - \mu_{i, j}) (r_{j, u} - \mu_{j, i})}{\sqrt{\sum_{u \in \mathcal{R}_{i, j}} (r_{i, u} - \mu_{i, j})^2} \sqrt{\sum_{u \in \mathcal{R}_{i, j}} (r_{j, u} - \mu_{j, i})^2}},
\end{displaymath}
where $\mathcal{R}_{i, j}$ is the set of all users who rated both $i$ and $j$, and
$\mu_{i, j}$ is the mean rating given to item $i$ by users who rated both $i$ and $j$.
If $\mathcal{R}_{i, j}$ is empty, then $\varrho_{i, j}$ is undefined. The
Pearson correlation coefficient $\varrho_{i, j}$ measures the tendency of
users to rate items $i$ and $j$ similarly. To avoid biased estimates in cases where
$n_{i, j} = \lvert\mathcal{R}_{i, j}\rvert$ is very small, we derive a new measure of similarity
\begin{displaymath}
	s_{i, j} = \frac{n_{i, j}}{n_{i, j} + \lambda} \cdot \varrho_{i, j}
\end{displaymath}
from $\varrho_{i, j}$ by shrinking towards zero \cite{KorenFactNgbr}. Here,
$\lambda \geq 0$ is a regularization parameter. Finally, we carry over these similarity
into distances by applying a logarithmic transformation:
\begin{displaymath}
	d_{i, j} = -\ln\left(\frac{1 + s_{i, j}}{2}\right).
\end{displaymath}
To derive a $d$-dimensional coordinate space in which items $i$ and $j$ approximately have distance $d_{i, j}$,
we use metric multidimensional scaling \cite{MDS}.
Since neighborhood-based recommendation methods are usually outperformed by factor models,
we expect our baseline coordinate space to be far inferior to those constructed using factor models.
We refer to our baseline model as MDS.

\section{EXPERIMENTS ON MOVIELENS\,10M}

We applied our approach to the MovieLens\,10M data set \footnote{\url{http://www.grouplens.org/node/73}},
which consists of about 10 million ratings collected by the online movie recommender service
MovieLens\footnote{\url{http://www.movielens.org}}. After postprocessing the original data (removing one non-existing movie,
merging several duplicate movie entries, and removing movies that received less than 20 ratings),
our new data set consists of 9{,}984{,}419 ratings of 8938 movies provided by 69878 users.
The ratings use a 10-point scale from 0.5 (worst) to 5 (best). Each user contributed at least 14 ratings.

Our analysis requires the genre information maintained by IMDb, so
we had to map each movie in the data set to its corresponding IMDb entry.
This task has been simplified a lot by the fact that all items in the MovieLens\,10M data set
are relatively well-known movies developed for cinema.\footnote{This is the reason why
we did not consider the Netflix data set. It consists of all kinds of
DVD titles, which often lack a clear correspondence in IMDb.}
We mapped about 8000 movies automatically by comparing titles and release years; the remaining
movies have been assigned manually or semi-automatically.

To avoid the problem of learning from very small samples for now, we did not use all 28 genres distinguished by IMDb.
Instead, we take only those genres into consideration that have been assigned to at least
5\% of all movies in our data set. Table\,\ref{genres} lists all remaining 13 genres and
their relative frequencies. On average, 2.3 genres have been assigned to each movie.

\begin{table}
	\centering
	\begin{tabular}{@{}lrclr@{}} \toprule
			Genre & \% & \hspace{5em} & Genre & \%\\ \midrule
			Action & 16.0 & & Horror & 10.1\\
			Adventure & 12.7 & & Mystery & 9.1\\
			Comedy & 38.2 & & Romance & 25.2\\
			Crime & 16.6 & & Sci-Fi & 8.6\\
			Drama & 54.6 & & Thriller & 24.2\\
			Family & 8.4 & & War & 5.2 \\
			Fantasy & 8.3\\ \bottomrule
		\end{tabular}
	\caption{Relative frequencies of genres.}
	\label{genres}
\end{table}

\subsection{Generating Coordinate Spaces}

We implemented each of the four coordinate extraction methods in MATLAB
and executed them on our rating data.

For SVD, $\delta$-SVD, and NNMF, we followed the literature
and used an optimization procedure based on gradient descent; to reduce computation time,
we applied the Hessian speedup proposed in \cite{RaikoSpeedup}. Adapting
the common methodology, we chose the regularization parameter $\lambda$ by cross-validation
such that the SSE is minimized on randomly chosen test sets. We ended up with
a value of $\lambda = 0.04$ for each of the three algorithms.

Since optimization by gradient descent is known to get stuck in local extrema
of the function to be minimized, we ran the three procedures at least three times,
each with different initial coordinates, which have been chosen randomly.
For each result, we computed the standardized solution pair as described in the previous
section. We found that the solutions generated by each extractor do not differ
significantly after standardization. This indicates that our coordinate spaces
match the unique solution of each optimization problem.

For our MDS procedure, we used the regularization constant $\lambda = 20$, which we determined
by adapting the recommendation Koren gave for the Netflix data set \cite{KorenFactNgbr}.
The coordinates have been generated by MATLAB's \texttt{mdscale} function using the
metric stress criterion. Since in our data set about 14 percent of all movie--movie pairs
had no raters in common, we treated the respective entries of the distance matrix as missing data.

To measure the effect of dimensionality, we generated three different coordinate spaces
with each extractor by varying the parameter $d$. We chose $d = 10$, $d = 50$, and $d = 100$.

\subsection{Applying the Classifiers}

In total, we used 14 different classifiers to evaluate each of the 12 coordinate spaces
with respect to each of the 13 genres.

We implemented the two support vector machine classifiers by soft-margin SVMs with parameters
$C = 4$ and (for SVM-RBF) $\gamma = 0.1$, which have been determined by cross-validation
to maximize classification accuracy.

Each of the four different kNN-classifiers will be applied to the data sets with three different
choices of $k$. To measure whether movies of the same genre tend to occur in larger groups,
we chose $k = 1$, $k = 3$, and $k = 9$. In the following, we will refer to these 12 classifiers
as $k$NN-Eucl, $k$NN-sEucl, $k$NN-scal, and $k$NN-cos.

To enable comparisons among classifiers and data sets, we generated 20 pairs of training and test sets,
each by randomly chosing 40\% of all movies for training and 10\% (of the remaining movies) for testing.
For each of the resulting 2184 combinations of coordinate spaces, classifiers and genres, we use the same
20 pairs of item sets for training and testing. In each case, we measured the classification accuracy.
All results reported below are averages over the 20 runs.

\subsection{Results}

Probably the most popular way of assessing a classifier's performance is measuring its accuracy, that is,
the fraction of test items which have been classified correctly. However, in our setting, this measure
is not very helpful. To see this, recall that the relative frequency of genres is very different in our
data set. For example, over half of all movies belong to the genre \emph{Drama,} but there are only about 5\% \emph{War} movies.
While attaining an accuracy of 95\% would be significant for the genre \emph{Drama,} it can easily be achieved
for the genre \emph{War} just by classifying any movie as \emph{non-War}. To enable comparisons
across genres, we propose to use a modified version of Cohen's kappa measure.

Any result of a binary classification task can be described by four numbers, which sum up to $1$:
the fraction of true positives ($\alpha_\text{tp}$), the fraction of false positives ($\alpha_\text{fp}$),
the fraction of false negatives ($\alpha_\text{fn}$), and the fraction of true negatives ($\alpha_\text{tn}$).
Accuracy is defined as $acc = \alpha_\text{tp} + \alpha_\text{tn}$. Moreover, the accuracy of a static majority-based
classifier (which always returns the label of the more frequent class) is
$acc_\text{maj} = \max\{\alpha_\text{tp} + \alpha_\text{fn}, \alpha_\text{fp} + \alpha_\text{tn}\}$.
We propose to use this kind of naive classifier for normalizing the accuracy and define
$\kappa = (acc - acc_\text{maj}) / (1 - acc_\text{maj})$. This measure expresses a classifier's relative
performance with respect to the majority-based classifier. If $acc = 1$ then $\kappa = 1$,
if $acc > acc_\text{maj}$, then $\kappa > 0$, if $acc = acc_\text{maj}$, then $\kappa = 0$, and if
$acc < acc_\text{maj}$, then $\kappa < 0$.

By measuring accuracy in terms of $\kappa$, we can average classification performance over different genres.
Tables\,\ref{kappas1}--\ref{kappas4} report the mean $\kappa$s over all 260 classification results obtained
for each combination of coordinate space and classifier type. All entries larger than 0.10 have been
marked in boldface. We can observe the following:

\begin{table}
	\centering
	\begin{tabular}{@{}rrrr@{}} \toprule
		& SVD-10 & SVD-50 & SVD-100\\ \midrule
		SVM-lin & 0.08 & \textbf{0.18} & \textbf{0.20}\\
		SVM-RBF & \textbf{0.15} & \textbf{0.23} & \textbf{0.25}\\
		1NN-Eucl & $-$0.24 & $-$0.21 & $-$0.19\\
		3NN-Eucl & 0.01 & 0.05 & 0.04\\
		9NN-Eucl & \textbf{0.12} & \textbf{0.16} & \textbf{0.14}\\
		1NN-sEucl & $-$0.25 & $-$0.27 & $-$0.31\\
		3NN-sEucl & 0.01 & 0.00 & $-$0.06\\
		9NN-sEucl & \textbf{0.12} & \textbf{0.12} & 0.04\\
		1NN-scal & $-$0.42 & $-$0.30 & $-$0.30\\
		3NN-scal & $-$0.16 & $-$0.03 & $-$0.03\\
		9NN-scal & 0.01 & \textbf{0.11} & \textbf{0.12}\\
		1NN-cos & $-$0.25 & $-$0.18 & $-$0.16\\
		3NN-cos & 0.00 & 0.06 & 0.06\\
		9NN-cos & \textbf{0.12} & \textbf{0.17} & \textbf{0.16}\\ \bottomrule
	\end{tabular}
	\caption{Kappas for coordinates generated by SVD.}
	\label{kappas1}
\end{table}

\begin{table}
	\centering
	\begin{tabular}{@{}rrrr@{}} \toprule
		& $\delta$-SVD-10 & $\delta$-SVD-50 & $\delta$-SVD-100\\ \midrule
		SVM-lin & 0.07 & \textbf{0.16} & \textbf{0.18}\\
		SVM-RBF & \textbf{0.13} & \textbf{0.20} & \textbf{0.23}\\
		1NN-Eucl & $-$0.26 & $-$0.26 & $-$0.26\\
		3NN-Eucl & $-$0.01 & 0.01 & $-$0.02\\
		9NN-Eucl & \textbf{0.11} & \textbf{0.12} & 0.08\\
		1NN-sEucl & $-$0.26 & $-$0.29 & $-$0.36\\
		3NN-sEucl & 0.00 & $-$0.03 & $-$0.11\\
		9NN-sEucl & \textbf{0.11} & 0.09 & $-$0.01\\
		1NN-scal & $-$0.41 & $-$0.28 & $-$0.22\\
		3NN-scal & $-$0.06 & 0.02 & 0.06\\
		9NN-scal & 0.05 & \textbf{0.13} & \textbf{0.16}\\
		1NN-cos & $-$0.26 & $-$0.19 & $-$0.16\\
		3NN-cos & 0.00 & 0.07 & 0.09\\
		9NN-cos & \textbf{0.12} & \textbf{0.18} & \textbf{0.19}\\ \bottomrule
	\end{tabular}
	\caption{Kappas for coordinates generated by $\delta$-SVD.}
	\label{kappas2}
\end{table}

\begin{table}
	\centering
	\begin{tabular}{@{}rrrr@{}} \toprule
		& NNMF-10 & NNMF-50 & NNMF-100\\ \midrule
		SVM-lin & 0.02 & 0.05 & \textbf{0.11}\\
		SVM-RBF & 0.02 & 0.09 & \textbf{0.14}\\
		1NN-Eucl & $-$0.56 & $-$0.47 & $-$0.41\\
		3NN-Eucl & $-$0.20 & $-$0.16 & $-$0.13\\
		9NN-Eucl & $-$0.02 & 0.01 & 0.02\\
		1NN-sEucl & $-$0.56 & $-$0.47 & $-$0.45\\
		3NN-sEucl & $-$0.20 & $-$0.16 & $-$0.16\\
		9NN-sEucl & $-$0.02 & 0.01 & 0.00\\
		1NN-scal & $-$0.37 & $-$0.34 & $-$0.34\\
		3NN-scal & $-$0.11 & $-$0.10 & $-$0.09\\
		9NN-scal & $-$0.02 & 0.00 & 0.02\\
		1NN-cos & $-$0.56 & $-$0.45 & $-$0.41\\
		3NN-cos & $-$0.20 & $-$0.15 & $-$0.13\\
		9NN-cos & $-$0.03 & 0.02 & 0.03\\ \bottomrule
	\end{tabular}
	\caption{Kappas for coordinates generated by NNMF.}
	\label{kappas3}
\end{table}

\begin{table}
	\centering
	\begin{tabular}{@{}rrrr@{}} \toprule
		& MDS-10 & MDS-50 & MDS-100\\ \midrule
		SVM-lin & $-$0.16 & \textbf{0.15} & \textbf{0.19}\\
		SVM-RBF & 0.03 & \textbf{0.16} & \textbf{0.17}\\
		1NN-Eucl & $-$0.29 & $-$0.19 & $-$0.18\\
		3NN-Eucl & $-$0.01 & 0.06 & 0.06\\
		9NN-Eucl & \textbf{0.13} & \textbf{0.18} & \textbf{0.18}\\
		1NN-sEucl & $-$0.29 & $-$0.23 & $-$0.29\\
		3NN-sEucl & $-$0.01 & 0.05 & $-$0.01\\
		9NN-sEucl & \textbf{0.13} & \textbf{0.17} & \textbf{0.12}\\
		1NN-scal & $-$0.29 & $-$0.19 & $-$0.18\\
		3NN-scal & $-$0.01 & 0.07 & 0.08\\
		9NN-scal & \textbf{0.12} & \textbf{0.18} & \textbf{0.18}\\
		1NN-cos & $-$0.28 & $-$0.18 & $-$0.16\\
		3NN-cos & 0.00 & 0.07 & 0.08\\
		9NN-cos & \textbf{0.13} & \textbf{0.19} & \textbf{0.19}\\ \bottomrule
	\end{tabular}
	\caption{Kappas for coordinates generated by MDS.}
	\label{kappas4}
\end{table}

\begin{itemize}
	\item
		The coordinate space derived by NNMF does not contain much helpful information about
		genres that can be exploited by our classifiers. The performance in all other spaces
		is significantly better.
	\item
		Except for NN-sEucl, classification performance generally improves with increasing dimensionality.
		However, the difference in performance between $d = 10$ and $d = 50$ is much larger
		than the one between $d = 50$ and $d = 100$. This indicates that our ordering of dimensions
		during standardization indeed captures some notion of relative importance. This is probably
		also the reason for NN-sEucl's decreasing performance with growing $d$; treating all dimensions
		equally seems to overweight information from dimensions at the end of the list.
	\item
		The SVM-RBF classifier slightly outperforms SVM-lin, but is comparable in
		performance to 9NN-Eucl, 9NN-scal, and 9NN-cos. This indicates
		that genres indeed tend to cluster in coordinate spaces, even with
		respect to different measures of distance.
	\item
		The NN-classifiers display bad performance for $k = 1$ and $k = 3$, which indicates
		that, although movies of the same genre roughly occur in clusters,
		each cluster usually also contains movies that do not have assigned the respective genre.
	\item
		In contrast to our expectations, the performance in coordinate spaces generated by factor models
		is comparable to the performance shown on our baseline coordinate space MDS.
\end{itemize}

Moreover, the results suggest that the performance of $k$NN-classifiers might even further increase for larger
values of $k$. To check this, we performed some preliminary tests with $k \approx 20$, but have not been able
to confirm this conjective.

We also investigated the influence of individual genres on classification performance;
as an example, the results for SVM-RBF are reported in Table\,\ref{kappasgen}. Entries larger than 0.20 have been
indicated. We can see that some genres, such as \emph{Horror} and \emph{Drama}, can clearly be identified by the classifier,
while others cannot. We have expected much better performance on clear-cut genres such as \emph{War.}

\begin{table}
	\centering
	\begin{tabular}{@{}rrrrr@{}} \toprule
		& SVD-100 & $\delta$-SVD-100 & NNMF-100 & MDS-100\\ \midrule
		Action & \textbf{0.34} & \textbf{0.31} & \textbf{0.22} & \textbf{0.22}\\
		Adventure & 0.13 & 0.12 & 0.08 & 0.00\\
		Comedy & \textbf{0.45} & \textbf{0.42} & \textbf{0.25} & \textbf{0.42}\\
		Crime & 0.08 & 0.06 & $-$0.01 & 0.00\\
		Drama & \textbf{0.47} & \textbf{0.43} & \textbf{0.37} & \textbf{0.44}\\
		Family & \textbf{0.43} & \textbf{0.46} & \textbf{0.31} & \textbf{0.34}\\
		Fantasy & 0.03 & 0.05 & 0.01 & 0.00\\
		Horror & \textbf{0.56} & \textbf{0.54} & \textbf{0.31} & \textbf{0.61}\\
		Mystery & 0.06 & 0.04 & $-$0.00 & 0.00\\
		Romance & 0.11 & 0.10 & $-$0.00 & 0.00\\
		Sci-Fi & \textbf{0.23} & 0.20 & 0.09 & 0.00\\
		Thriller & \textbf{0.31} & \textbf{0.27} & 0.14 & 0.15\\
		War & 0.05 & 0.06 & $-$0.00 & 0.00\\ \bottomrule
	\end{tabular}
	\caption{Kappas for SVM-RBF by genre.}
	\label{kappasgen}
\end{table}

In summary, these preliminary experiments suggest that the coordinate spaces derived by SVD, $\delta$-SVD,
and MDS indeed contain some significant semantic information about the represented movies.
However, the situation is by far not as clear as claimed by the literature.

\section{CONCLUSION AND OUTLOOK}

In the current paper, we presented a general methodology for systematically analyzing
whether coordinate spaces generated from factor models contain semantic information,
as it is commonly claimed. We applied our approach to the MovieLens\,10M data set
and found initial evidence for this claim.

Our results encourage us to follow this line of research in several ways.
First, we would like to investigate whether our results also carry over to
more advanced and complex factor models, which have been proposed very recently \cite{KorenSVD++, KorenFactNgbr}.
It would also interesting to see what more traditional methods such as multidimensional
scaling can contribute to the problem of feature extraction from rating data,
since our results indicate that these methods can sucessfully be modified
for use in our new setting.



\bibliography{factorsemantics_mpref}

\end{document}